\documentclass[a4paper,twoside]{article}

\usepackage{epsfig}
\usepackage{subcaption}
\usepackage{calc}
\usepackage{amssymb}
\usepackage{amstext}
\usepackage{amsmath}
\usepackage{amsthm}
\usepackage{multicol}
\usepackage{pslatex}
\usepackage{apalike}
\usepackage{SCITEPRESS}     

\usepackage{hyperref}
\usepackage[T1]{fontenc}
\usepackage[utf8]{inputenc}

\usepackage{xcolor}

\begin{document}

\title{Developing a Successful Bomberman Agent}

\author{
\authorname{Dominik Kowalczyk, Jakub Kowalski, Hubert Obrzut,\\ Micha{\l} Maras, Szymon Kosakowski, Rados{\l}aw Miernik}
\affiliation{Institute of Computer Science, University of Wroc{\l}aw, Wroc{\l}aw, Poland}
\email{xkowal99@gmail.com, jko@cs.uni.wroc.pl, skoczek133@gmail.com,\\ mmaras1999@gmail.com, szymon.kosu@gmail.com, radoslaw.miernik@cs.uni.wroc.pl}
}


\abstract{In this paper, we study AI approaches to successfully play a 2--4 players, full information, Bomberman variant published on the CodinGame platform. We compare the behavior of three search algorithms: Monte Carlo Tree Search, Rolling Horizon Evolution, and Beam Search. We present various enhancements leading to improve the agents' strength that concern search, opponent prediction, game state evaluation, and game engine encoding.
Our top agent variant is based on a Beam Search with low-level bit-based state representation and evaluation function heavy relying on pruning unpromising states based on simulation-based estimation of survival. It reached the top one position among the 2,300 AI agents submitted on the CodinGame arena.}

\onecolumn \maketitle \normalsize \setcounter{footnote}{0} \vfill

\section{\uppercase{Introduction}}
\label{sec:introduction}

Games were always used as testbeds for Artificial Intelligence, being a motivation for the development of new methods and showcase of the current advancements.
From the classic boardgames like Backgammon \cite{tesauro1994td}, Checkers \cite{schaeffer2007checkers}, Chess \cite{Campbell2002Deep}, Go \cite{Silver2016Mastering}, to video games like Pac-Man \cite{rohlfshagen2017pacman}, Mario \cite{togelius20102009}, Hearthstone \cite{dockhorn2019introducing}, Atari games \cite{mnih2015human}, and Starcraft \cite{Ontanon2013ASurvey,vinyals2019grandmaster}.
These approaches resulted in the development and advancements of many algorithms such as alpha-beta pruning, temporal difference learning, Monte Carlo Tree Search, Rolling Horizon Evolutionary Algorithm, Deep Neural Networks, etc. \cite{Russell2020AIM}.

At the same time, thousands of commercially published video games were developed relying solely on ``less advanced'' techniques to control believable agents and player opponents -- such as BFS, DFS, A$^*$, Finite State Machines, Decision Trees, Behavior Trees, and Hierarchical Task Networks \cite{rabin2013game,millington2009artificial}.
Still, there are games where the speed and relative simplicity of these approaches allow them to solve challenges posed by academic research -- just to mention Mario AI Competition beaten by Robin Baumgarten's A$^*$ \cite{togelius20102009}.

In this paper, we study AI approaches to successfully play a Bomberman-like game \cite{Bomberman}, called \emph{Hypersonic}.
We present our comparison of three different algorithmic approaches (Monte Carlo Tree Search, Rolling Horizon Evolution, and Beam Search), various constructions of the evaluation function, and the influence of the game engine efficiency on the obtained results.
Apart from internal testing, we have validated our agent on the online arena provided by the CodinGame platform, containing currently about 2,300 AI agents.

The paper is structured as follows. In the next section, we briefly introduce game tree search algorithms used, Bomberman in AI research, and the CodinGame platform.
The following section describes the \emph{Hypersonic} game in more detail.
In Section~\ref{sec:engine} we concisely compare the efficiency of our two engine representations.
Next, we present the components we used to evaluate the game state.
Section~\ref{sec:algorithms} describes our algorithm implementations.
The following section contains the results of the cross-algorithms experiments and the description of our submissions on the CodinGame arena.
Finally, we conclude in Section~\ref{sec:conclusion}.

\vfill

\section{\uppercase{Background}}

\subsection{Algorithms}

\subsubsection{Beam Search}\label{sec:beamintro}
Beam Search \footnote{\url{https://en.wikipedia.org/wiki/Beam_search}} is a classic greedy algorithm merging the breadth-first search approach with a heuristic evaluation function to prune nodes on each level of the tree.
In the standard implementation, at each level we store at most \emph{beam width} states, and only those nodes are further expanded. Their descendants are evaluated using a heuristic function and only the best \emph{beam width} are stored for the next round.

Beam Search is considered a good choice for games that have limited interaction with the opponent(s), so the planning part is more important than the exact opponent prediction. 
At the cost of losing the optimality, the algorithm allows us to easily trade branching factor for increased search depth -- thus gain an advantage on future planning. 
For this base variant, multiple extensions have been proposed, e.g. Beam Stack Search that makes the algorithm optimal \cite{zhou2005beam}.
Other standard enhancements, such as transposition tables\footnote{\url{https://www.chessprogramming.org/Transposition_Table}}, can be successfully applied as well.

\subsubsection{Monte Carlo Tree Search}

Monte Carlo Tree Search (MCTS) is a stochastic, anytime, asymmetric, knowledge-free search algorithm popular in a wide range of games and non-games problems \cite{Browne2012ASurvey,swiechowski2021monte}.
It has been especially successful in the domains of computer Go \cite{Silver2016Mastering} and various challenges of General Game Playing \cite{perez2019general,finnsson2010learning}.

The algorithm gradually builds a partial game tree combining random simulations of the game with a selection of nodes (usually based on the UCB1 formula \cite{kocsis2006bandit}) that guides search towards promising paths.

Conceptually, the algorithm is knowledge-free. It does not require heuristic knowledge about the game, and the statistical data is gathered using true game rewards obtained in the terminal states.
However in practice, for many games, it is impossible (or unprofitable) to perform full game simulations, and it is more beneficial to use a heuristic evaluation function to estimate the quality of game states at a certain depth.

\subsubsection{Rolling Horizon Evolutionary Algorithm}

Rolling Horizon Evolutionary Algorithm (RHEA)  approaches game tree search as a planning problem and encodes sequences of actions as genomes \cite{perez2013rolling}. It uses standard evolutionary operators (recombination and mutation) to sparsely search for good sequences in the domain of possible agent actions up to some depth.
RHEA has been successfully used in multiple domains, outperforming MCTS in games with continuous space \cite{samothrakis2014rolling}, multi-action games \cite{justesen2016online}, and also in General Video Game AI \cite{gaina2020rolling}.

A search performed by RHEA is less ``methodical'', and in contrast to MCTS, rather than on accumulated statistics, RHEA
relies on (hopefully existing) similarities between good sequences of actions.
The algorithm is highly parameterizable and versatile, depending on the choice of representation and operators. As it does not rely on statistical significance, it often requires less computations than MCTS to find a good plan to follow.
On the downside, using operators may lead to illegal action sequences; thus, usually, some repair procedure is required.
Also, opponent handling is less standardized -- but also more customizable -- with multiple approaches that can be used \cite{Liu2016Rolling}.
RHEA is often combined with heuristic functions that evaluate the game states' quality after applying actions from the genome.

\subsection{Pommerman}

Pommerman is a variant of Bomberman developed as a benchmark for AI agents, featuring competitive and cooperation skills in multi-agent partially observable scenario \cite{resnick2018pommerman}. 
The game can be played in three modes: free-for-all (FFA), team, and team with communication. 
Also, the observations received by the agents can be limited by their vision range -- a number of tiles around the players with the visible content (everything else is hidden).
The vision can be set to infinity, resulting in a fully observable game variant.
Pommerman, in a partially observable team variant without communication, has been used as a NeurIPS 2018 competition.

In this competition, top agents were based on the MCTS algorithm, including the winner and the agent that placed third, described in \cite{osogami2019real}.
A comparison in playing strength of some simple agents against MCTS, BFS, and Flat Monte Carlo has been shown in \cite{zhou2018hybrid}.
More comprehensive analysis focusing on statistical forward planning algorithms, namely MCTS and RHEA, can be found in
\cite{perez2019analysis}.

Additionally, various deep learning methods were applied to learn how to play the game offline: 
e.g., Relevance Graphs obtained by a self-attention mechanism \cite{malysheva2018deep}, or continual learning to train the population of advantage-actor-critic agents \cite{peng2018continual}.
The runner-up of the learning track of the NeurIPS competition is described in 
\cite{gao2019skynet}. Its network is trained using Proximal Policy Optimization and improved with reward shaping, curriculum learning, and action pruning.

Before the NeurIPS competition, a smaller one was launched, focusing on the FFA mode \cite{resnick2018pommerman}. The winner used Finite State Machine Tree Search, while the runner-up was a handcrafted, rule-based agent.

\subsection{CodinGame}

CodinGame\footnote{\url{https://www.codingame.com}} is a challenge-based learning/coding platform created in 2015 that contains multiple types of activities, mainly related to AI programming. 
These include a long list of algorithmic problems to solve (as usual for such type of platform \cite{combefis2016learning}) and competitive programming, which requires writing an agent that directly competes against programs written by other players. 
Currently, the platform supports 27 programming languages that may be used to solve all of the available tasks.

Most of the proposed activities are game-like and interactive, i.e., the user program sends actions and gets responses from the system with the observations that can be used to compute the next action.
Also, tasks on CodinGame usually have pleasant visualizations, making them more appealing and easier to debug.
Thus, the site generally receives positive feedback as a tool for education \cite{butt2016students}.

A few times per year, CodinGame organizes contests based on newly released games. Two competitions launched in 2020 gathered about 7,000 and 5,000 programmers respectively.
Available problems and games are developed not only by the CodinGame team, but also by the community.

Some of them were created for research purposes, e.g., \textit{TotalBotWar} described in \cite{estaben2020totalbotwar}. The \textit{Strategy Card Game AI Competition}\footnote{\url{https://legendsofcodeandmagic.com}} running since 2019 with IEEE COG and CEC conferences is based on a past CodinGame contest. 
Related research focuses on choosing the right deck to play \cite{Kowalski2020EvolutionaryApproach,vieira2020drafting} as well as optimizing heuristic policies to select proper actions \cite{montoliu2020efficient}.

\section{\uppercase{Hypersonic}}

\begin{figure}
  \centering
\includegraphics[width=\linewidth]{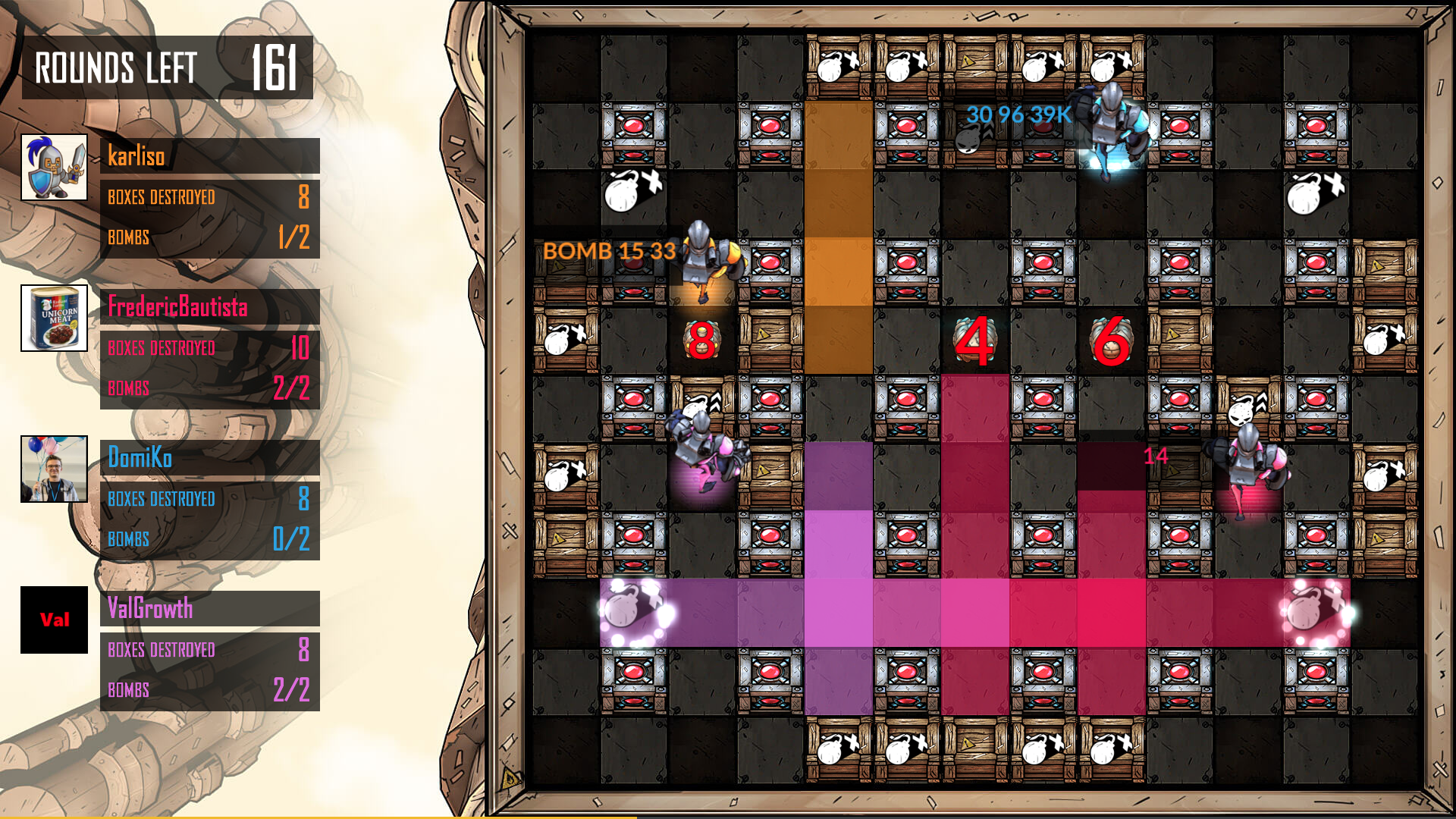}
  \caption{
    A visualization of the Hypersonic game.
  }
  \label{fig:hypersonic}
\end{figure}

\begin{figure}
  \centering
\includegraphics[width=\linewidth]{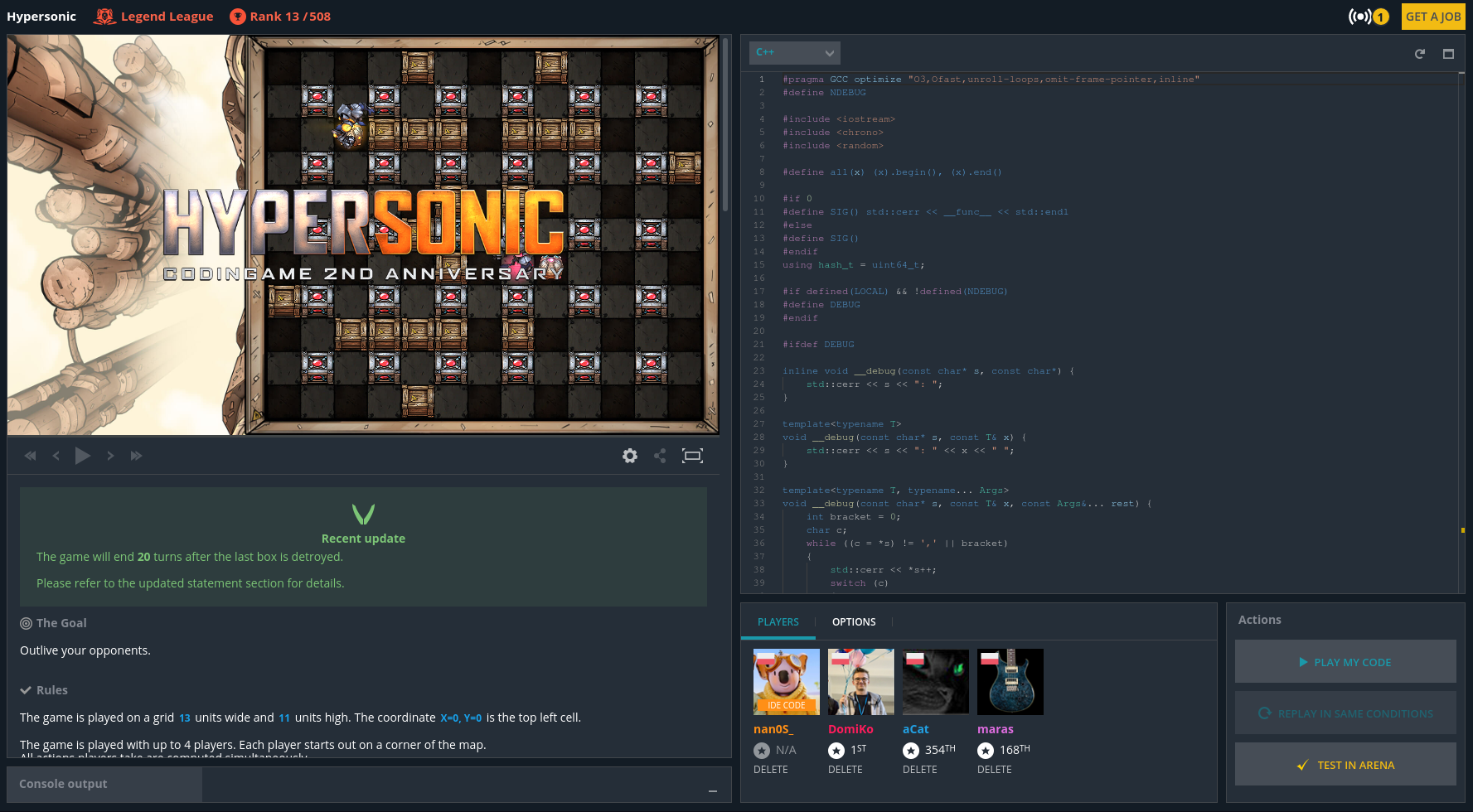}
  \caption{
    Playing Hypersonic in the CodinGame online IDE. Top left -- game visualization, bottom left -- information about current turn and debug, top right -- agent source code, bottom right -- opponent setup for testing.
  }
  \label{fig:hypersonicide}
\end{figure}

\emph{Hypersonic} is an adaptation of Bomberman \cite{Bomberman} as a programming game available on the CodinGame platform (Fig.~\ref{fig:hypersonic}).
It was initially public for a competition held in September 2016. 
The game attracted over 2,700 participants and was very well-received by the community.
After the contest ended, Hypersonic became available as always ongoing multiplayer activity\footnote{\url{https://www.codingame.com/multiplayer/bot-programming/hypersonic}}. 
Thus, everyone can write his agent using an online CodinGame IDE (Fig.~\ref{fig:hypersonicide}) and test it on a public arena that currently contains about 2,300 players.
About 500 of them belong to the highest, \emph{Legend} league, which requires writing a program that is better than a predefined agent called \emph{boss}.\\

\noindent \textbf{Game Rules.} The game is played by 2 to 4 players on 13$\times$11 grids. Grid cells may be floors (passable by the players), walls, or boxes (impassable).
Walls are always placed in the same way (every odd cell in both directions), while boxes are placed randomly but symmetrically across the grid. 
Boxes can be destroyed by bombs (in contrast to walls, which are indestructible).
They may contain items inside, and destroying a box with an item makes this item appear on the map.

Players start in corners of the map. Possible actions are to place a bomb on the current cell (or not) and  independently to move towards a given cell (which on an atomic level is equal to move left, right, up, down, or stay in place), so there are up to 10 possible actions per player. Players cannot move on a cell containing a wall, box, or bomb. All players perform their moves simultaneously, and multiple players can occupy the same cell. If a player moves on a cell containing an (unboxed) item, they collects it (several players can collect the same item if they arrived simultaneously).

Placed bombs explode after 8 turns, creating blasts horizontally and vertically up to \emph{range} cells. If a blast encounters another bomb its explosion is automatically triggered, causing a chain reaction. Such explosions eliminate players, destroy boxes, and remove unboxed items. Explosions are computed and applied before players' actions.

Initially, players may place one bomb with a range of 3. Items that may drop from boxes can give a player additional bombs or increase his bombs' range.

Players have full knowledge about the game state and the moves of the opponents.
This includes a full visibility range and knowing the content of undestroyed boxes (if any).

Finally, players are ranked by the order of elimination (the winner is the last one standing).
Ties are resolved by counting the number of boxes destroyed by the player's bombs -- a higher number wins.
If all boxes are gone, the game ends automatically after 20 turns.
The turn limit is set to 200, but for the top tier bots' games usually take between 90--120 turns for two players, and 70--100 turns for four-player games.

The time limits required by the CodinGame arena are 1000 ms for the first turn and 100 ms for each of the following turns. 
The communication protocol is based on standard input/output streams -- each turn player reads the information about the actual game state and outputs a single line with an action to perform. 

The general rules of Hypersonic are roughly comparable to Pommerman free-for-all mode played in a fully observable variant. 
However, note that certain small differences have a significant impact on how the game is actually played, e.g. varying number of players, no bomb kicking action, destroyed boxes deciding on player ranking.

\vspace{-.5\baselineskip}
\section{\uppercase{ENGINE}}
\label{sec:engine}
\vspace{-.5\baselineskip}

In order to allow game tree search, we require a game engine. It has to efficiently simulate the course of the game, and support the following operations:
\begin{itemize}
    \item Computing the legal actions for a given player, which is used both in computing our agent's best actions and simulating enemies.
    \item Keeping the current game statistics, which are used in the state evaluation; mainly for our agent (more in Section~\ref{sec:stateeval}).
    \item Computing the next game state given a set of actions of all the players (as all players perform their actions simultaneously).
\end{itemize}

Usually, the game engine is also responsible for detecting terminal states and returning the players' rewards. In our approach, however, we treat the game as potentially infinite and simulate the future without taking into account the earlier endgame. This is due to the fact that after all boxes in Hypersonic are gone, then the secondary score is set and the only goal becomes survival. Moreover, most of the top bots become passive at this stage avoiding risky actions.

Since the speed of the engine has a significant impact on the performance of the algorithms, everything was implemented in the C++ language. We have developed two versions of the engine: one is a very straightforward implementation of the game rules, while the other relies heavily on bitwise operations and additional preprocessing.

The bitwise version reduces the complexity of many engine functions, including blast propagation, to constant time (instead of depending on the number of items and the board size) thus it gives a significant boost in performance.
To show the scale of the improvement, we presented a comparison of simulation speed between both engines in Table~\ref{table:engine_comparison}.

\vspace{\baselineskip}
\renewcommand{\arraystretch}{1.05}
\begin{table}[h]
  \centering
  \caption{
    \centering
    The total number of actions performed by a random agent during  $500ms$ starting from example midgame situation with $2$, $3$, and $4$ players, resetting every $15$ actions or death.
  }
  \begin{small}
    \begin{tabular}{r|c|c|c}
    Number of players            & 2     & 3    & 4 \\ \hline
          \texttt{Naive engine}             & $90k$ & $79k$ & $80k$ \\
          \texttt{Bitwise engine}           & $1.45m$ & $1.3m$ & $1.28m$ \\
    \end{tabular}
  \end{small}
  \label{table:engine_comparison}
\end{table}

\section{\uppercase{STATE EVALUATION}}\label{sec:stateeval}

Besides standard engine functions related to the game simulation, we implemented a few evaluation-related helper functions that are calculated along the way. They typically require a couple of additional fields (increasing the size of the stored game state) but are very helpful in determining a more precise state evaluation. Overall, it translates to a higher quality of the agent. 

The majority of these helper functions refer to the ``survivability problem'' as this is a very common issue in Hypersonic, especially when playing against more players. Determining if a state is survivable for our agent, allows us to prune bad states early and concentrate on more promising ones.

These helper state evaluation functions are:
\begin{itemize}
    \item \texttt{estimated\_bombs(gamestate)} -- estimates score for destroyed boxes. For each box destroyed during the simulation, we add $\gamma^d$, where $d$ is the number of turns in the future when the bomb will explode. 
    \item \texttt{is\_survivable(gamestate, player)} -- returns true if the player can survive, assuming the opponents will not place any new bombs. It is checked using an eight-turns deep BFS.

    \item \texttt{can\_kill(gamestate, player, enemy)} -- returns true if the enemy can successfully trap and eliminate the player within the next two rounds. This is calculated through an exhaustive search.
\end{itemize}

Then, we may calculate the evaluation function, which is used to estimate the quality of the final (not necessarily terminal in terms of the game rules) state reached by the simulation.
The value of the function for the game state $\mathit{state}$ is the sum of:

\begin{enumerate}
\item $1$ point for each destroyed box,
\item $0.9 \times min(5, range) + 0.4 \times range$ points
\item $3.4 \times min(2, extra\_bombs)\ + $ \\ 
    $1.7 \times min(4, extra\_bombs)\ + $\\
    $0.7 \times extra\_bombs$ points

\item $\mathit{estimated\_bombs}(\mathit{state})$ with $\gamma=0.95$,
\item $0.05 \times \mathit{distance\_to\_other\_players}$,
\item if there are more than $20$ boxes remaining:\\ $-0.04 \times \mathit{distance\_to\_center}$, \\otherwise $-0.1 \times \mathit{average\_distance\_to\_boxes}$

\item $-1000$ points if the player is dead.

\end{enumerate}

\section{\uppercase{IMPLEMENTED ALGORITHMS}}\label{sec:algorithms}

We have tried a few conceptually different approaches -- evolutionary (RHEA), statistical (MCTS), and greedy (Beam Search). Each approach has its roadblocks that we had to overcome.
From the preliminary carried out experiments, it seems that the best choice for Hypersonic is based on a greedy approach -- Beam Search agent performed considerably better than both MCTS and RHEA agents.
This aligns with the experience of the majority of other contestants, as the best bots in CodinGame leaderboard use a modified Beam Search algorithm. 

For this reason, the development of the Beam Search version received most of our attention, and we tuned this algorithm carefully to work well against the opponents on the CodinGame arena. 

All algorithm parameters were chosen experimentally, based on local testing, as well as the performance on the CodinGame arena.\\

\noindent \textbf{Opponent Prediction.}
Each of the algorithms applies a similar strategy to predict the opponents' behavior.
At the beginning of every turn, we run the same algorithm as for our agent to simulate each of the opponents.
These opponents plan their actions with the assumption that every other agent does not move.
Then, we apply the search for our agent with the enemies following the approximated action sequences instead of standing still.

Application of this general opponent prediction schema slightly differs between algorithms.
MCTS and RHEA simulate each opponent for $10ms$, and Beam Search allocates $15ms$ in total for all the opponents.
MCTS and RHEA need more time to find any reasonable sequence of moves, when $5ms$ for each player seemed enough for Beam Search.

\subsection{MCTS}
\vspace{-.5\baselineskip}

There are multiple variants of MCTS that seem eligible for the Bomberman-type game, most notably Single-Player Monte-Carlo Tree Search (SP MCTS) \cite{sp-mcts} and Monte Carlo Tree Search for Simultaneous Move Games \cite{MCTSforSimultaneousMoveGames}. The second one requires heavy computational power -- each player has up to $10$ moves during a turn, so every
node might have up to $10^4$ children. Even with the highly optimized engine, we were not able to achieve a reasonable amount of iterations, hence we abandoned this method.

In order to use the Single-Player MCTS algorithm, one can completely ignore enemy players -- given an initial game state, the opponents never change their position nor
plant any bombs. In that way, the number of children for each node in the search tree is significantly reduced, but the agent is unable to predict any incoming traps
from the enemy players and detect which boxes will most likely be destroyed by the enemies in the following turns. 

As previously described, to address the issue with predicting enemies, we run SP MCTS for each one of them and generate an approximated sequence of their future moves.
Additionally, to avoid any possible traps, for each possible agent's action in the initial state we run an exhaustive search for two turns in the future and check if any opponent is able to eliminate the agent. If so, we omit that action in the search tree.

According to the rules, each game may have up to $200$ turns. In $100$ ms the SP MCTS tree usually reaches depth around $12$ for the agent and $9$ for the opponents. Since the approximated action sequences for the enemies are shorter and the enemies may not even follow them, very deep simulations are not reliable. Thus, we limit the depth of the random simulations.
If during the simulation the game did not end, we use a heuristic function to evaluate the value of the state.
In addition, we completely ignore the $200$ round limit in the rules -- the game ends only if all players but one are eliminated.\\

\noindent \textbf{Evaluation.}
Defining a good heuristic function for the state evaluation poses a great challenge. Because the state values for the MCTS need to be well balanced, we were not able to use the same state evaluation as described in the previous section. Most of the random simulations end with the agent getting eliminated, so the penalty for the agent's death cannot be too high -- otherwise, the MCTS will prefer states with the least probability of death following a random policy, and the agent will be highly discouraged to plant any bombs. We reward the agent for each box it destroyed. We apply exponential decay to the points achieved from destroying the boxes so that the agent is encouraged to do it as soon as possible.

Because the depth of the simulations is rather small, the agent may have a hard time finding sequences of actions to destroy boxes that are far away. To obtain better results, the agent is rewarded for being close to the remaining boxes on the game board. The experimentally chosen parameters for the MCTS and the heuristic function were as follows:

\begin{itemize}
    \item In the selection phase of the MCTS the classic UCT method was used, with $c$ constant set to $1$.
    \item In each node we store two values: the mean and the maximum of the results of random simulations.
    \item When choosing the action to perform in the actual play, the one with the highest maximum is taken.
    \item Depth of the random simulations is $15$.
    \item For each round $r$ of the simulation, $\mathit{reward}_r$ is equal to the number of boxes the agent destroyed and an additional point for each extra bomb power-up until the agent has at least $4$ bombs.
    \item The value of the final simulation state is $0$ if the agent died during a simulation. Otherwise, it is calculated as
    \[\frac{1}{400}\times(\sum_{r=1}^{15}\mathit{reward}_r\times 50 \times  \gamma^{r} + 200 - \mathit{box\_dist}),\]
    where $\gamma = 0.98$, and $\mathit{box\_dist}$ is the sum of the Manhattan distances between the agent and each box on the game board.
\end{itemize}

\subsection{RHEA}

The main algorithm schema looks like this:
\begin{enumerate}
    \item Maintain $\mathit{population\_size}$ of chromosomes, each is a sequence of actions of length $\mathit{chromosome\_length}$.
    \item Select parents based on fitness and use crossover operators to create $\mathit{offspring\_size}$ children.
    \item Mutate children with $\mathit{mutation\_probability}$.
    \item Create a new population using replacement operator.
\end{enumerate}

One challenge in order for RHEA to work is to properly evaluate an individual. It has to be noticed that not every sequence of moves is a valid one, e.g. we could have a wall ahead, we could have no bombs in the inventory, or another bomb blocking a path. This becomes especially true when we apply evolutionary operators which are generally game rules agnostic. The other problem is that performing an action could lead us to death, which could be easily avoidable.  Thus, in our solution, we ignore actions that are invalid or result in the agent's death. Also, additional effort was made to discourage individuals which, although end up living, are not in a survivable state. So the evaluation score of the individual is the game state evaluation after all valid actions are performed, reduced by punishment value. The further in the future the first moment in which we are in a not survivable state the smaller the punishment. If we end up in a survivable state the punishment is zero.\\

\noindent The following evolutionary operators were used:
\begin{enumerate}
    \item \textbf{crossover:} we have considered standard one-point crossover and uniform crossover operators. Performed tests indicated that the one-point crossover was much better. One explanation could be that the one-point crossover is very natural when it comes to problems related to pathfinding (which our game has a lot in common).
    \item \textbf{mutation:} we replace every gene of every child with a random value with probability $\mathit{mutation\_probability}$.
    \item \textbf{selection:} parents are selected using roulette wheel method based on their fitness.
    \item \textbf{replacement:} we have considered $(\mu, \lambda)$ and $(\mu + \lambda)$ replacement with full elitism. It seems that $(\mu + \lambda)$ replacement is much more efficient as it keeps previously found good solutions.
\end{enumerate}

The parameter values we have used are $\mathit{mutation\_probability} =0.5$, $\mathit{population\_size} =50$, $\mathit{offspring\_size} = 50$, $\mathit{chromosome\_length} = 17$.

\subsection{Beam Search}
\begin{figure}[h!]
  \centering

  \includegraphics[trim={10px 5px 40px 40px},clip,width=.97\linewidth]{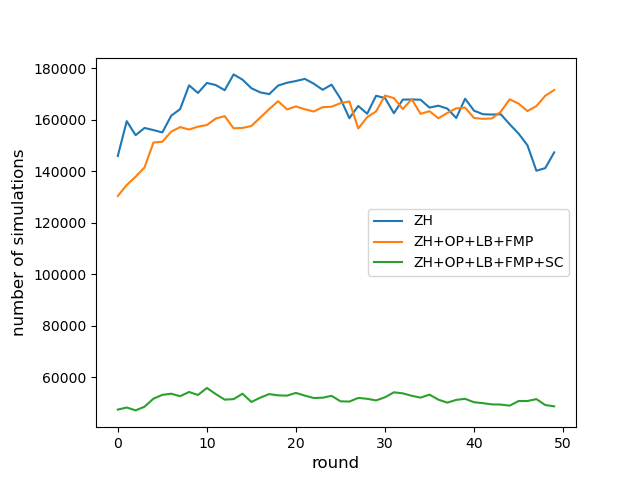}
  \caption{Number of simulations per game round for particular sets of features (100ms per round).}
  \label{fig:Beam Search_iters}
\end{figure}

\begin{figure}[h!]
  \centering
  \includegraphics[trim={10px 5px 40px 40px},clip,width=.97\linewidth]{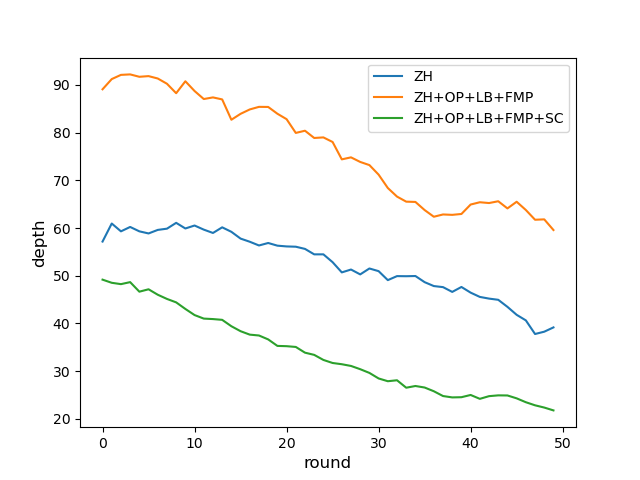}
  \caption{Depth reached by the Beam Search per game round for particular sets of features (100ms per round). 
  Note that result of the \texttt{ZH}-only version is due to the twofold increase of the beam width.}
  \label{fig:Beam Search_depth}
\end{figure}

We extended a vanilla Beam Search schema by adding a few enhancements. Notably:
\begin{itemize}
    \item \texttt{Zobrist hashing (ZH).}  At each level we remove duplicates by computing a hash function of every state using Zobrist hashing.
    \item \texttt{Opponent prediction (OP).} We approximate the enemies' future actions as previously described.
    \item \texttt{Local beams (LB).} We group states by the player position and restrict that for each position we can store at most $\mathit{local\_beam\_width}$ best states (according to our evaluation function). This allows to maintain diversity and improve the quality of the search while preserving the greediness of the algorithm. Still, the global restriction of storing at most $\mathit{beam\_width}$ states at each level remains.
    \item \texttt{First move pruning (FMP).} We prune all the actions available in the initial state which are not survivable. Additionally, if there is at least one action that leads to enemy death, we prune all the others. Those computations are done using \texttt{is\_survivable} and \texttt{can\_kill} functions. 
    
    On some rare occasions it may be beneficial for the agent to kill both the enemy and itself -- according to the rules, if every player is eliminated from the game, the one that destroyed more boxes wins. Whenever the agent is able to commit suicide while also killing the last enemy, we calculate the number of points each player is able to obtain and if the win is certain, we force the agent to apply this action.
    \item \texttt{Survivability checking (SC).} We highly discourage states in the beam that are not survivable, by decreasing their scores. Survivability, similarly as in first move pruning, is detected with $\mathit{is\_survivable}$ function. Although this imposes great computational costs (see Figures~\ref{fig:Beam Search_iters} and \ref{fig:Beam Search_depth}), it allows us to get rid of the bad states quickly and focus more on the high-quality ones.
\end{itemize}

The parameter values we have used are $\mathit{beam\_width} = 500$ and $\mathit{local\_beam\_width} = 12$.

We have tested agent versions with each feature \texttt{OP}, \texttt{LB}, \texttt{FMP} against a vanilla version (\texttt{ZH} only, $\mathit{beam\_width}$ increased to 1000).
The comparison of the above improvements is presented in Table~\ref{table:BeamSearch}. 

\renewcommand{\arraystretch}{1}
\begin{table}[h!]
  \centering
  \caption{
    \centering
    Influence of Beam Search improvements on agent's strength compared to the vanilla (i.e. containing only \texttt{ZH}) version. The first row  provides a baseline (\texttt{ZH} vs \texttt{ZH}). 500 games per pair.
  }
  \begin{small}
    \begin{tabular}{l|c|c}
       Enhancement     & WIN     & LOSE \\
      \hline
          \texttt{ZH}               & 19.00\% & 19.00\% \\
          \texttt{ZH+OP}            & 44.80\% & 19.60\% \\
          \texttt{ZH+LB}            & 45.20\% & 36.20\% \\
          \texttt{ZH+FMP}           & 45.00\% & 29.20\% \\
          \texttt{ZH+OP+LB+FMP}     & 57.60\% & 22.60\% \\
          \texttt{ZH+OP+LB+FMP+SC}  & 59.40\% & 23.40\% \\
    \end{tabular}
  \end{small}
  \label{table:BeamSearch}
\end{table}

We did not test a single \texttt{SC} feature, as survivability checking does not work standalone, only when combined with the other improvements.
Moreover, it is a very special feature given its computational overhead. 
We had to turn \texttt{SC} off in the other algorithms, as it was too costly to be beneficial there.
However in Beam Search, given the speed achieved by the bit-based engine, computations above a certain threshold do not transfer to the improved player's strength meaningfully, making some computational power ``free''.

Thus, even sacrificing half of the simulations performed is worth obtaining a small improvement on the win ratio, especially that this feature seems more beneficial on the CodinGame arena than in the local algorithm comparison.

\section{\uppercase{EXPERIMENTS}}\label{sec:experiments}

We have performed a direct comparison of each algorithm in its best-performing version we encoded.
The experiments were run using the \emph{cg-brutaltester} tool\footnote{\url{https://github.com/dreignier/cg-brutaltester}}, under the standard limitation of 100ms per game round. 
Table~\ref{table:algcomparison} shows the results of one versus one games, while in Table~\ref{table:algcomparison3} we put results of three-player games with all agents playing against each other.

As we can see, the advantage of the Beam Search version is tremendous. However, the performance on the three-player games is noticeably smaller, as these games are more unpredictable.
Although the Beam Search approach was put the most consideration in our development, even a plain version without enhancements easily scores above 80\% against the other two. Comparing MCTS and RHEA, the former performs clearly better in both duels and three-player experiments.

\renewcommand{\arraystretch}{1.05}
\begin{table}[h]
  \centering
  \caption{
    \centering
    Win percentages for each algorithm in 1 vs 1 setting. A single cell shows the win ratio of the row agent versus the column agent. Each pair of agents was evaluated on $500$ matches.
  }
  \begin{small}
    \begin{tabular}{r|c|c|c}
                                & \texttt{\ MCTS\ } & \texttt{\ RHEA\ }  & \texttt{Beam Search}         \\
      \hline
          \texttt{MCTS}         & ---         & 67.80\%  & 2.20\%         \\
          \texttt{RHEA}         & 22.40\%     & ---      & 1\%            \\
          \texttt{Beam Search}  & 96.60\%     & 99\%     & ---            \\
    \end{tabular}  
  \end{small}
  \label{table:algcomparison}
\end{table}

\renewcommand{\arraystretch}{1.05}
\begin{table}[h]
  \centering
  \caption{
    \centering
    Win percentages of the algorithms in 1 vs 1 vs 1 setting. A single cell shows in how many games the row agent obtained a higher score than the column agent. Agents were evaluated on $1000$ matches.
  }
  \begin{small}
    \begin{tabular}{r|c|c|c}
                                & \texttt{\ MCTS\ }    & \texttt{\ RHEA\ }      & \texttt{Beam Search} \\
      \hline
          \texttt{MCTS}         & ---               & 68.70\%           & 10.30\%   \\
          \texttt{RHEA}         & 23.80\%           & ---               & 3.70\%   \\
          \texttt{Beam Serach}  & 84.20\%           & 94.90\%           & ---       \\
    \end{tabular}  
  \end{small}
  \label{table:algcomparison3}
\end{table}

\subsection{Agents behavior}
The Beam Search approach is definitely the most successful and shows stable ``inhuman'' performance. The complexity and depth of the search make analyzing the games by humans very hard.

Thus, it is not possible to tell the reason a bot won or lost in each game.
Other algorithms are easier to examine, as they sometimes show clear tendencies to some unexpected behaviors.

For example, the MCTS agent feels a strong need to place bombs every turn when it seems safe and it does not see their clear use in the future. We have to admit that we miserably failed to unlearn him this quirk. 
On a serious note, despite the opponent's prediction, the MCTS agent still from time to time can be trapped by the enemies bombs blocking his way out.
Also, given how the values computed by the evaluation function influence the node selection in the UCT, tuning the proper weights is much harder than for the other two approaches.

On the other hand, RHEA suffers mostly from highly influential nondeterminism. With unlucky initial sequences, it may waste many generations until producing something useful (and each generation is costly as any chromosome modification requires reevaluation) or get stuck at some local optimum that is very hard to get out of. 
This algorithm has a tendency to often produce sequences with oscillating actions that are not so easy to prune because of the unknown in advance bomb blast possibilities.
Also, because of the illegality of the produced action sequences, the actual search horizon is often significantly smaller than the declared chromosome length.

\subsection{CodinGame Ranking}
\begin{figure}[th]
  \centering
\includegraphics[width=\linewidth]{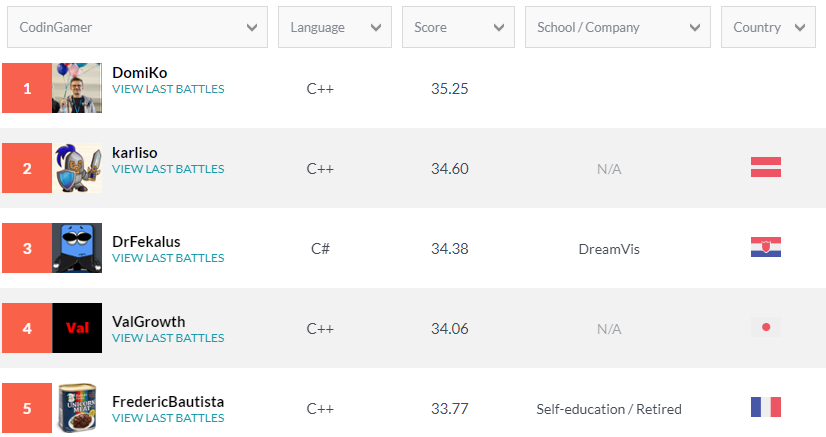}
  \caption{
    A screenshot from the CodinGame Hypersonic leaderboard (taken 29.11.2021), with our Beam Search algorithm variant on the first position.
  }
  \label{fig:leaderboard}
\end{figure}

\begin{figure}[th]
  \centering
\includegraphics[width=\linewidth]{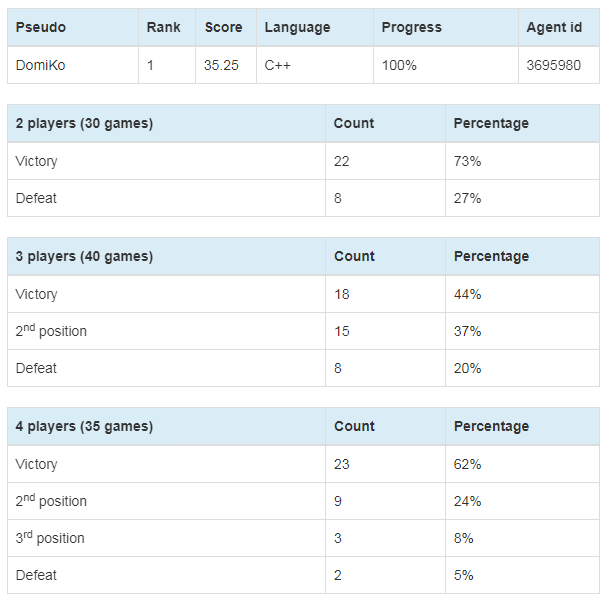}
  \caption{
    A screenshot from the CGStats showing detailed win-rates depending on the number of players.
  }
  \label{fig:cgstats}
\end{figure}

To present our work in a wider context, we show in Figure~\ref{fig:leaderboard} the results from the CodinGame Hypersonic leaderboard\footnote{\url{https://www.codingame.com/multiplayer/bot-programming/hypersonic/leaderboard}}, with our best-performing Beam Search agent ranked in the first place. Our agents based on MCTS and RHEA were placed at positions $183$ and $204$ respectively. 
The agent scores are computed using the True Skill algorithm.  
To show specific win-rates of the agent, depending on the number of players, in Figure~\ref{fig:cgstats} we show an excerpt of the statistics provided by the CGStats tool\footnote{\url{http://cgstats.magusgeek.com/}}. 

The source code of our agents is not publicly available. The CodinGame code of behavior forbids posting full source codes to avoid copying it by other members of the community. (Just to note that in Hypersonic many places in high legend are occupied by the clones of one agent that broke this rule.) In the future, we plan to release partial code that will contain as much useful information as possible, without providing an easy way to be copy-pasted as a working agent.

\section{\uppercase{CONCLUSION}}\label{sec:conclusion}

We presented our investigations over creating an AI playing Hypersonic, a CodinGame adaptation of Bomberman, that finally lead us to develop a top-performing agent.
Along the way, we tested few approaches, including various algorithms, engine codification, and components of the evaluation function.
In this paper, we show some of our findings that we think might be useful for people applying described algorithms for similar games.

Finally, we focused on a Beam Search with low-level bit-based state representation and evaluation function heavily relying on pruning unpromising states using simulation-based estimation of survival.
Our best version of the algorithm was able to reach the first position among the $2,300$ AI agents posted on the CodinGame Hypersonic arena.

As future work, we plan to perform a more exhaustive parameter search to make sure the agent is optimized against the other top leaderboard contestants.
In the extended version, we want to describe in detail the bitwise state representation and how to achieve constant-time blast propagation, as well as provide more interesting figures regarding the behavior of each of the algorithms tested and the influence of particular elements of our evaluation function.

\bibliographystyle{apalike}
{\small
\bibliography{bibliography}}

\end{document}